\documentclass[final]{cvpr}

\usepackage{times}
\usepackage{epsfig}

\usepackage{graphicx}
\usepackage{amsmath}
\usepackage{bm}
\usepackage{amssymb} 
\usepackage{subfigure}

\usepackage{multirow}
\usepackage{algorithm}
\usepackage{algorithmic}
\usepackage{xcolor}
\usepackage[pagebackref=true,breaklinks=true,colorlinks,bookmarks=false]{hyperref}

\def\cvprPaperID{11115} 
\def\confYear{CVPR 2021}
\newcommand{\R}[1]{\textcolor{red}{#1}}
\newcommand{\bI}{\boldsymbol{I}}  
\newcommand{\bt}{\boldsymbol{t}} 
\newcommand{\bc}{\boldsymbol{c}} 
\newcommand{\bS}{\boldsymbol{S}} 
\newcommand{\bG}{\boldsymbol{G}} 
\newcommand{\cS}{\mathcal{S}} 
\newcommand{\cW}{\mathcal{W}} 
\newcommand{\btheta}{\boldsymbol{\theta}} 
\newcommand{\ba}{\boldsymbol{a}} 
\newcommand{\vect}[1]{\boldsymbol{#1}}
\definecolor{mygray}{gray}{0.5}

\begin{document}

\title{WSSOD: A New Pipeline for Weakly- and Semi-Supervised Object Detection}

\author{
    Shijie Fang$^1$ \and Yuhang Cao$^2$ \and Xinjiang Wang$^1$ \and Kai Chen$^2$ \and Dahua Lin$^2$ \and Wayne Zhang$^1$ \\
    $^1$ SenseTime Research \\
    $^2$ CUHK - SenseTime Joint Lab, The Chinese University of Hong Kong\\
{\tt\small \{fangshijie, wangxinjiang, wayne.zhang\}@sensetime.com} \\
 \tt\small \{yhcao6, chenkaidev\}@gmail.com \\ 
 \tt\small dhlin@ie.cuhk.edu.hk \\ 
}

\maketitle
\thispagestyle{empty}


\begin{abstract}
The performance of object detection, to a great extent, depends on the availability of large annotated datasets.
To alleviate the annotation cost, the research community has explored a number of ways to exploit unlabeled or weakly labeled data. However, such efforts have met with limited success so far.
In this work, we revisit the problem with a pragmatic standpoint, trying to explore a new balance between detection performance and annotation cost by jointly exploiting fully and weakly annotated data.
Specifically, we propose a weakly- and semi-supervised object detection framework (WSSOD), which involves a two-stage learning procedure.
An agent detector is first trained on a joint dataset and then used to predict pseudo bounding boxes on weakly-annotated images.
The underlying assumptions in the current as well as common semi-supervised pipelines are also carefully examined under a unified EM formulation.
On top of this framework, weakly-supervised loss (WSL), label attention and random pseudo-label sampling (RPS) strategies are introduced to relax these assumptions, bringing additional improvement on the efficacy of the detection pipeline.
The proposed framework demonstrates remarkable performance on PASCAL-VOC and MSCOCO benchmark, achieving a high performance comparable to those obtained in fully-supervised settings, with only one third of the annotations.
\end{abstract}

\section{Introduction}
\label{sec:intro}

With recent advances in deep learning, object detectors such as Faster-RCNN\cite{ren2015faster}, 
RetinaNet\cite{lin2017focal} and FCOS\cite{tian2019fcos} have made great impact.
Their success partially attributes to the the large-scale datasets.
However, building such a dataset with accurate bounding box annotations is time consuming and laborious. 
The reliance of object detection on detailed annotations poses a great challenge for industrial applications, where only scarce annotations may be available.

There have been attempts to alleviate this problem, e.g., weakly-supervised object detection (WSOD) and semi-supervised object detection (SSOD).
WSOD~\cite{bilen2016weakly, tang2018pcl, lin2020object, shen2019category, tang2017multiple} avoids the high cost of labeling bounding boxes with weakly-annotated data.
Only image-level labels (i.e. categories of objects in image) are utilized to train detectors.
Most of these methods are based on multi-instance learning (MIL)\cite{dietterich1997solving}.
SSOD\cite{jeong2020interpolation, tang2019learning, chen2020temporal, sohn2020simple, jeong2019consistency} makes use of a few fully-annotated data (i.e., with both category labels and bounding box coordinates) as well as a larger amount of unlabeled data.
Training based on consistency regularization\cite{jeong2020interpolation, tang2019learning, jeong2019consistency} or pseudo labels\cite{wang2018weakly,sohn2020simple} are two popular frameworks in SSOD. 
However, the performance of both settings still fall far behind the fully-supervised counterpart. The absence of box-wise annotations in WSOD is a bottleneck for predicting accurate spatial bounding boxes. Thus the state-of-the-art WSOD method only achieves 10.2\% mAP on MSCOCO dataset \cite{shen2019category}.
As for the SSOD algorithms, the utilization of unlabeled data is still unsatisfactory since the noise caused by incorrect gradients may be accumulated and hurt the convergence. Therefore, there still exists a large performance gap between semi-supervised and fully-supervised learning. 

\begin{figure}[!t]
   \subfigure[Fully-supervised]{
      \begin{minipage}{0.44\linewidth}
         \includegraphics[width=1.6in]{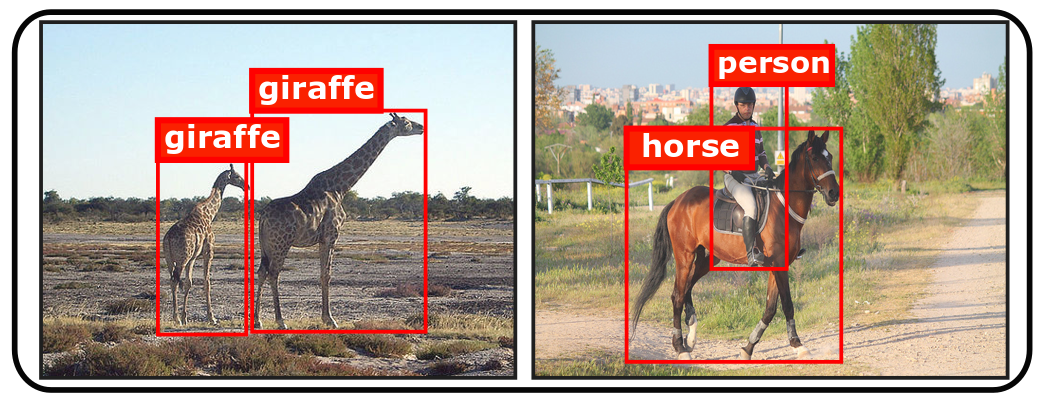}
      \end{minipage}
   }
   \subfigure[Weakly-supervised]{
      \begin{minipage}{0.44\linewidth}
         \includegraphics[width=1.6in]{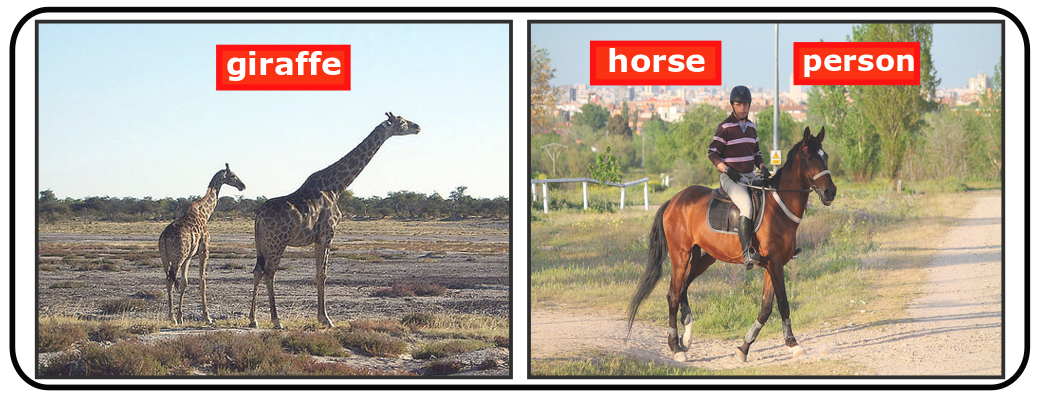}
      \end{minipage}
   }

   \subfigure[Semi-supervised]{
      \begin{minipage}{0.44\linewidth}
         \includegraphics[width=1.6in]{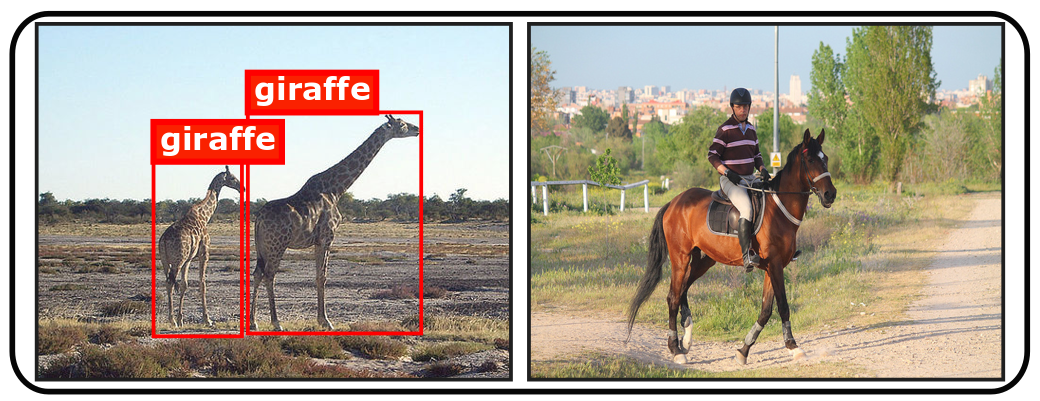}
      \end{minipage}
   }
   \subfigure[Weakly- and Semi-supervised]{
      \begin{minipage}{0.44\linewidth}
         \includegraphics[width=1.6in]{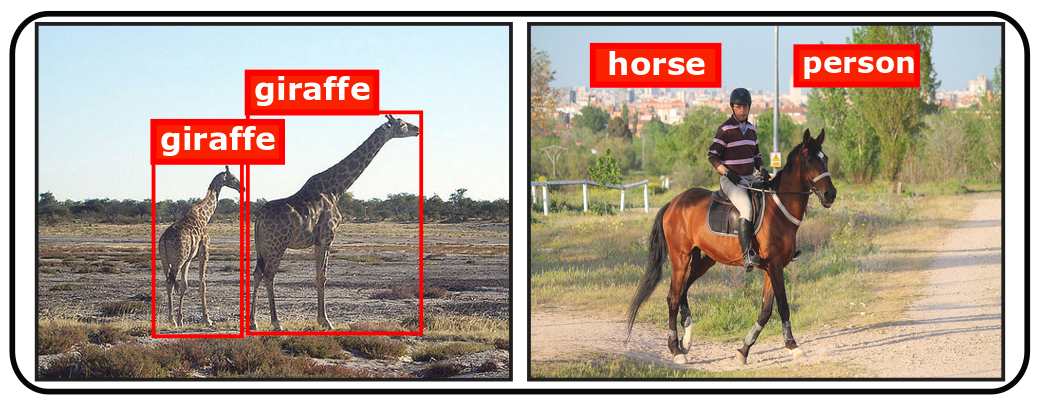}
      \end{minipage}
   }
   \centering
   \label{fig: different_setting_detection}
   \caption{Different settings of object detection.}
\end{figure}

Although remarkable achievements have been made in WSOD and SSOD, the non-negligible performance gap still hind them from real applications.
Considering the limitation of those explorations, a natural question raises: \textit{Is there a better trade-off between annotation cost and performance?}
In this paper, we propose to train detectors in a weakly- and semi- supervised manner. Under this setting, the detector is trained with a few fully-annotated data as well as abundant weakly-annotated data.
The idea stems from the imbalanced labor in annotating images for classification and detection tasks. 
For instance, it only takes a few seconds to annotate an image with category labels but can take minutes to annotate all bounding boxes, especially in crowded scenes.Therefore it is possible to obtain lots of weakly labeled data with a relatively small expense. 

In this study, we jointly adopt a small number of fully annotated samples and a large number of weakly-annotated images to train an object detector, forming a Weakly- and Semi-Supervised Object Detection (WSSOD) pipeline.
The fully-annotated samples serves as important anchors while the weakly-annotated samples help increase the generalization ability of models.
In this way, our method shrinks the gap with fully-annotated baselines and achieves high detection performance while maintaining an affordable annotation cost. 

The proposed WSSOD pipeline is rooted in an Expectation-Maximization (EM) view over the hidden variables, which are the unknown bounding box annotations in weakly-supervised data. 
Specifically, the designed pipeline is a single-circle EM with two stages.
In the first stage, an agent detector is trained on both fully and weakly labeled data, to yield a better initialization and more accurate estimate of the posterior probability of the hidden variables. 
The trained model is later used to predict bounding boxes and class labels on weakly labeled data, which is called pseudo label. 
In the second stage, we use fully-annotated data as well as pseudo-labeled data to train a target detector, so as to estimate and maximize the joint probability of both fully- and weakly-annotated data.

This pipeline shares slight similarity with FixMatch~\cite{sohn2020fixmatch} in semi-supervised classification.
We also noticed that some previous works~\cite{chen2020temporal, sohn2020simple} on object detection also utilized a similar pseudo-label based method with multiple stages. 

However, WSSOD differs from them on the following aspects. 
For one thing, the performance of agent detector has the most direct impact on the quality of pseudo label, thus further determines power of target detector. 
Unlike \cite{chen2020temporal, sohn2020simple} which only considered fully-annotated samples in stage 1, we also utilized the weakly-annotated data to boost the performance of agent detector. 
For another, within the framework of EM, we are able to explain the initiatives and drawbacks of various designs in the original pipeline. We also propose a label attention module to achieve a better estimation of the posterior probability of the hidden bounding boxes in weakly-annotated data. 
Moreover, an adaptive pseudo-label generation module is proposed to relax the assumptions when conducting EM, which addresses the ever-present problem to determine the threshold while generating pseudo labels. 
In contrast, precedent studies like~\cite{sohn2020simple} manually tuned the threshold (from $0.9$ to $0.5$) according to the performance of agent detector, which is resource consuming and unreliable.

In summary, the main contributions of our work are summarized as follows.

\begin{enumerate}
   \item We propose the WSSOD framework to train detectors in a weakly- and semi-supervised manner, which takes both the annotation cost and performance into account. 
   \item The current pipeline is designed and analyzed under the EM framework. In order to improve the posterior
   probability of the unknown bounding boxes in the weakly-annotated data, a novel attention module and loss function are employed in WSSOD.
   \item We list the major assumptions in common pipelines under the EM framework, and one major assumption is relaxed by the proposed adaptive pseudo label generation module, which also reduces the number of key hyper parameters and improves the quality of pseudo labeling.
   \item Results on the challenging MSCOCO 2017 dataset demonstrates the effectiveness of our method. Specifically, we achieve 36.1\% mAP using Faster-RCNN with only 30\% fully-labeled data, which performs comparably with the fully-supervised setting.
\end{enumerate}


\section{Related Work}
\label{sec:related}

\noindent
\textbf{Object Detection}
is a fundamental task in the field of computer vision, which has made great progress in recent years. 
The contemporary detectors can be categorized as two paradigms, two-stage detectors \cite{girshick2014rich,girshick2015fast,ren2015faster,lin2017feature,he2017mask,cai18cascadercnn} which first generate region proposals which are then classified and refined in the second stage; single-stage detectors \cite{redmon2017yolo9000,liu2016ssd,lin2017focal,tian2019fcos} removes the process of proposal generation but directly make predictions on top of the predefined anchor boxes. 
Though both of two paradigms are thoroughly studied in the circumstance where sufficient and fulled labeled data are given, weakly supervised and semi-supervised situations are less explored for object detections and the performance is still not satisfactory. In this paper, we try to explore a balance between detection performance and label cost by utilizing a small part of weakly image-level labels for a better semi-supervised learning. 

\noindent
\textbf{Weakly Supervised Object Detection (WSOD)}
is a challenging problem that aims to eliminate the need of the tight instance-level annotation but only gives image-level labels. 
Many studies \cite{gokberk2014multi, bilen2016weakly,song2014weakly} formulated WSOD as a Multiple Instance Learning (MIL) \cite{dietterich1997solving} problem and adopt an alternative pipeline to iteratively train the detector and infer the instance label. To alleviate the non-convex problem of MIL strategy, various initialization \cite{song2014weakly,deselaers2010localizing} and regulation methods \cite{diba2017weakly,song2014weakly,song2014learning} were proposed to prevent it from sticking at the local minimum. 
Recently, \cite{bilen2016weakly} proposed a two-stream network WSDDN to simultaneously perform region selection and classification. The region level scores from these two streams are then element-wise multiplied and transformed to image-level scores by summing over all regions.
Its subsequent works \cite{kantorov2016contextlocnet} which combines with context information and \cite{tang2017multiple} which combine with multi-stage refinement further push the performance. 
However, their works both relies on the traditional Selective Search \cite{uijlings2013selective} which is outdated. We leverage it with the objectness score predicted by Region Network and incorporate it to the latest detectors.

\noindent
\textbf{Semi-Supervised Learning (SSL)}
studies the scenario where annotated data is sparse. Two representative methods for SSL in image classification are consistency regularization and pseudo labeling. 
Consistency regularization, like \cite{xie2020self,sohn2020simple,devries2017improved,rasmus2015semi}, aims to regularize the network prediction when presenting a image and its augmented one. 
Pseudo labeling, like \cite{lee2013pseudo,xie2020self,pham2019semi,bachman2014learning}, where a teacher model is firstly trained on the labeled data and then used to make prediction on unlabeled data as pseudo labels.
The success of SSL for image classification inspires some recent researches in object detection. 
CSD \cite{jeong2019consistency} explores consistency regularization by enforcing the detector make consistent prediction on a image and its horizontally flipped one. 
Proposal learning \cite{tang2020proposal} adds noise to the proposal features instead of the raw image for better noise-robust proposals feature predictions. 
Noisy Student \cite{xie2020self} adopts the student-teacher pseudo labeling pipeline and ensembles extra classifier for eliminating the noise introduced by box mining phase. 
Recently, \cite{sohn2020simple} proposes a SSL framework STAC for object detection. In this paper, we follow this framework and adopt several improvements to the phases of teacher model training and pseudo label generation to improve the overall performance.

\section{Methodology}
\label{sec:methodology}
\subsection{Notation}
Let $\bI \in \mathbb{R}^{H\times W\times 3}$ denotes an image, $\bc \in \{0, 1\}^C$ represent the weakly-annotated multilabels of the image, $C$ be the total number of foreground categories. In the following theoretical derivation, $C=1$ is assumed for better clarity without loss of generalizations and therefore $\bc$ degrades into a scalar $c$. $\bt = (\tilde{\bt}_1, \tilde{\bt}_2, ...)$ is a full annotation of an image where each $\tilde{\bt}=(c, \vect{x})$ is a tuple representing a foreground instance $c=1$ with coordinate $\vect{x}$. We denote by $\mathcal{S}$ the index set of the fully-annotated images and by $\mathcal{W}$ the index set of the weakly-annotated images. $\btheta$ is the parameters of a neural network which takes $\bI$ as input and is able to output its box predictions. There are also $N$ proposals (or anchors) in an image to descretize the infinite-sized sliding windows in object detection, and the each proposal is denoted as $\ba$. We also slightly abuse the symbol $\bt = (\tilde{\bt}_1, \tilde{\bt}_2, ...\tilde{\bt}_N)$ to denote the model-generated bounding boxes for each proposal. In this case $\tilde{{\bt}}=(c=0, \varnothing)$ is allowed and denotes a background prediction. 

\subsection{Background: Expectation-Maximization}
The objective of weakly- and semi-supervised object detection is to maximize the joint probability of supervised and weakly-supervised data. 
\begin{equation}
\label{eq:total_p}
P = \prod_{i \in \cS} P(\bt_i|\bI_i; \btheta)\prod_{i\in \cW } P(c_i |\bI_i; \btheta).
\end{equation}
The optimization of $P(\bt_i|\bI_i; \btheta)$ is a well-studied problem and it's log-probability can be decomposed into 
\begin{equation}
\label{eq:log_s}
\begin{split}
\log{P(\bt|\bI; \btheta)} = &\sum_{j=1}^N {\log P(\tilde{\bt}_{s(j)}|\bI, \ba_j)} \\
                          = &\sum_{j \in \mathcal{A}_P}{\log P(\vect{x}_{s(j)}|\bI, \ba_j)} \\
                          & + \sum_{j \in \mathcal{A}}{\log P(c_{s(j)}|\bI, \ba_j)},
\end{split}
\end{equation}
where $\mathcal{A}$ and $\mathcal{A}_P$ denote the set of total and positive proposals. For common supervised object detection pipelines, the loss design can be viewed as the maximization of the above log-likelihood.
Assume the coordinates of each bounding box are independent and follow a Laplacian distribution and classification labels follow a Bernoulli distribution, then we yield the familiar combination of (smooth) L1 loss and cross-entropy loss for each term in Eq. \ref{eq:log_s}, 
\begin{equation}
\begin{split}
\log P(\tilde{\bt}_{s(j)}|\bI, \ba_j) &= \log P(c|\bI, \ba_j) + \log P(\vect{x}_{s(j)} |\bI, \ba_j) \\
&= - \left( \lambda_1 CE(c, \vect{p}) + \lambda_2 L1(\vect{x}, \vect{x_s} \right).
\end{split}
\end{equation}

As for the weakly-supervised probability, since $P(c=1|\bt) = 1$ if for each of the image-level label, there is at least one instance belonging to this category,  $P(c=1 |\bI; \btheta)$ can be simplified as
\begin{equation}
\begin{split}
P(c=1 |\bI; \btheta) &= \sum_{\bt}P(c=1, \bt |\bI; \btheta) \\ 
&= \sum_{\bt}P(\bt | \bI;c ; \btheta)P(c=1|\bt) \\
&= \sum_{\bt \in \mathcal{B}}P(\bt | \bI; c; \btheta), 
\end{split}
\end{equation}
where $\mathcal{B}$ is the set of proposal assignments that satisfy $P(c=1|\bt) = 1$. Note that each $\bt=(\tilde{\bt}_1, \tilde{\bt}_2, ..., \tilde{\bt}_N)$ actually represents one possible combination of the output from all proposals, and there are in total $2^N-1$ elements in $\mathcal{B}$. For example, $((c=1, \vect{x})_1, (c=0, \varnothing)_2, ..., (c=0, \varnothing)_N)$ and $((c=0, \varnothing)_1, (c=1, \vect{x})_2, , ..., (c=0, \varnothing)_N)$ are both valid elements in $\mathcal{B}$.  

The above equation involves the probability of hidden variables $\bt$, and EM is a common method to optimize it.
The EM algorithm over both the supervised and weakly-supervised data is the estimation and maximization of the following property 

\begin{equation}
\label{eq:Q1}
Q = \sum_{i \in \mathcal{S}}\log{P(\bt_i|\bI_i; \btheta)} + \sum_{i\in\mathcal{W}}{Q(c_i, \bI_i)},
\end{equation}
where
\begin{equation}
\label{eq:Q2}
Q(c, \bI) = \sum_{\bt \in \mathcal{B}}P(\bt | \bI; c; \btheta')\log P(\bt | \bI; \btheta),
\end{equation}
with $P(\bt | \bI; \btheta)$ expressed as Eq. \ref{eq:log_s}, and $\btheta'$ is the model parameter of the last EM iteration. 

\subsection{Pipeline}
\label{subsec:pipeline}
\begin{figure*}
	\centering
	\includegraphics[width=0.85\textwidth]{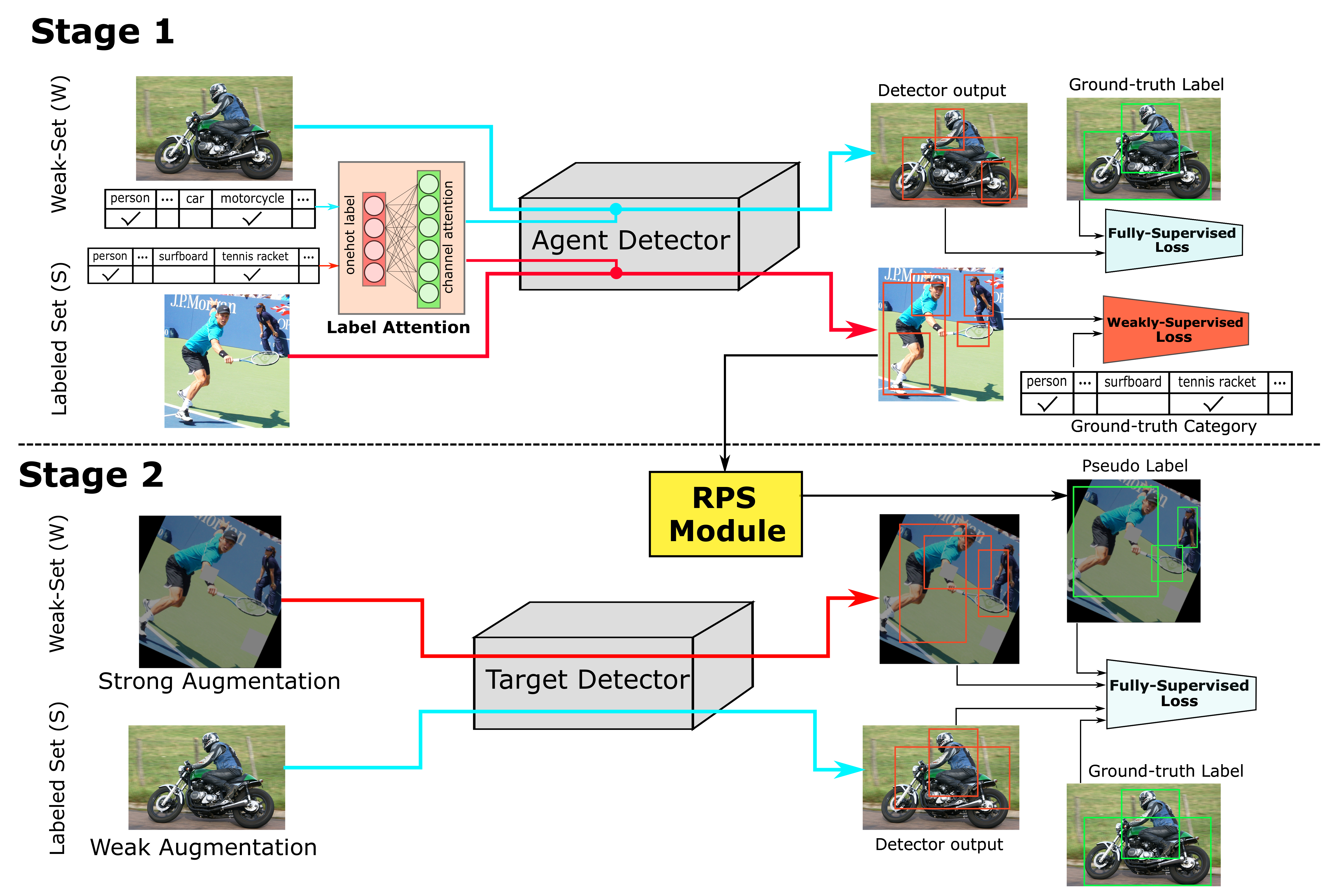}
	\caption{The pipeline design for WSSOD. The cyan curved arrow indicates the flow of fully-annotated data while the red one indicates weakly-annotated data. }
	\label{fig:pipeline}
\end{figure*}

Based on the EM framework, we design the pipeline as shown in Fig. \ref{fig:pipeline}. It is a two-stage iterative process. But in reality, only a single circle is conducted, constrained by the common training budget for fair comparison. Stage1 trains an agent detector and resembles the initialization of $\btheta'$ in Eq. \ref{eq:Q2} and tries to yield an accurate $P(\bt | \bI; c; \btheta')$. The instantiation of $P(\bt | \bI; c; \btheta')$ is realized by the pseudo-label generation. That is, the predicted bounding boxes of weakly-annotated images are samples from the agent detector and fed into Stage2 along with fully-annotated images as the target detector's training data. Stage2 performs both the E-step and M-step in the sense that the target detector first evaluates the loss (\emph{i.e.} $-Q$ in Eq. \ref{eq:Q1}) brought by the pseudo label and then optimize it through back-propagation.
For the training of weakly-annotated set in Stage2, we borrow the idea from semi-supervised learning \cite{sohn2020fixmatch, sohn2020simple} that apply strong augmentation $T$ only on weakly-labeled data and multiple a scalar $\lambda_u$ to the loss of them. That is to say, the Eq. \ref{eq:Q1} now becomes 
\begin{equation}
  \label{eq:Q3}
  Q = \sum_{i \in \mathcal{S}}\log{P(\bt_i|\bI_i; \btheta)} + \lambda_u \sum_{i\in\mathcal{W}}{Q(c_i, T(\bI_i))}.
\end{equation}
The brought $T$ can be treated as disturbance and $\lambda_u$ is a scale factor, which will not affect the optimization objective. 
So we still use the original notations for convenience.

Therefore, under the EM framework, our target now becomes how to effectively evaluate $Q$ as well as how to obtain a good agent model and further a more accurate estimation of $P(\bt | \bI; c; \btheta')$. 
\subsection{Stage1: towards an accurate $P(\bt | \bI; c; \btheta')$}
The major target in Stage1 is to obtain a good initialization of the agent detector, $\btheta'$, which is then used to yield a good $P(\bt | \bI; c; \btheta')$ as pseudo label for the target detector in the second stage. 
The training process also involves both $\cS$ and $\cW$ to yield a better $\btheta'$. 
The optimization for labeled set $\cS$ is identical to Eq. \ref{eq:log_s}, and in order to make use of the weakly-annotated set $\cW$, we borrow the idea in WSDDN \cite{bilen2016weakly} and propose a WSL loss.

\textbf{Weakly-supervised loss (WSL)}. The image-label probability for weakly-supervised images can be formulated as

\begin{equation}
\label{eq:wsl}
\begin{split}
P(c=1 |\bI; \btheta') & \sim \max_j{P(c_j=1 | \bI; \ba_j; \btheta')} \\ 
                      & \sim \mathrm{Softmax}_j{P(c_j=1 | \bI; \ba_j; \btheta')}, 
\end{split}
\end{equation}
where $j$ is the index of proposal $\ba_j$. The equation implies that maximizing the probability of an image label is converted to the maximization of the upper bound of the label probability among all proposals. The hard max function is then softened using $\mathrm{Softmax}$, over the classification score of all proposals. Despite the multiple assumptions involved during the derivation of Eq. \ref{eq:wsl}, as will be later unveiled in Sec. \ref{sec:stage2}, it still serves as an effective tool to leverage $\cW$ and therefore leads to a better initialization of $\btheta'$. 

As for object detectors with region proposal network (RPN), the  $P(c=1 |\bI; \btheta')$ is divided into the product of a prior (RPN score) and its posterior (bbox score),
\begin{equation}
\label{eq:wsl2}
\begin{split}
P(c^k=1 |\bI; \btheta') \sim & \sum_j [\mathrm{Softmax}_j{P_1(c_j=1 | \bI; \ba_j; \btheta')}  \\
                           & \cdot \mathrm{Softmax}_k{P_2(c_j^k=1 | \bI; \ba_j; \btheta')}], 
\end{split}
\end{equation}
where $P_1$ and $P_2$ denotes the score in RPN and bbox head, respectively, $j$ is the index of proposals and $k$ is the index of classes. 

The cross entropy between $P(c=1 |\bI; \btheta')$ and the given label in $\cW$ then leads to the weakly-supervised loss (WSL) adopted in this study.
In implementation, only $K$ (\emph{e.g.} $K=512$) proposals are sampled in the RPN stage to calculate their $\mathrm{Softmax}$ w.r.t. their objectness score, to reduce the influence of overwhelming simple background proposals. 
This implementation is different from that used in WSDDN \cite{bilen2016weakly} , which relies on the traditional selective search\cite{uijlings2013selective} and use an extra stream to predict the detection score.

\textbf{Label Attention}. The target of agent model in the EM framework is to provide an accurate estimation of $P(\bt | \bI; c; \btheta')$, instead of $P(\bt | \bI; \btheta')$ used in only semi-supervised settings. This offers us a great advantage when designing the agent model as it able to explicitly leverage the image label of $\cW$ in both \textbf{train} and \textbf{test} phase, since the test phase is also conducted on the train dataset to generate pseudo labels. This is also different from the target detector in Stage2, which should be designed to be unaware of any image annotations during test phase. 

Therefore, the small difference inspires us to design an explicit label attention module as displayed in Fig. \ref{fig:pipeline}. The input image label is directly encoded into a one-hot vector $\bc$, followed by an FC layer to generate the attention map to be fused with the feature map of the agent detector. 

$$\vect{F}(\bI) \leftarrow \textrm{Sigmoid}\left( \textrm{FC}(\bc) \right) \odot \vect{F}(\bI), $$
where $\vect{F}$ denotes an intermediate featuremap to be multiplied with the attention map from the image label over the featuremap channels.

\subsection{Stage2: towards an effective estimation of $Q$}
\label{sec:stage2}
Estimating $Q(c=1, \bI)$ in Eq. \ref{eq:Q2} is a super challenging task as it requires the summation over all possible proposal assignments, which for each foreground class requires computation complexity $\sim O(2^N-1)$, which means that as long as there are at least one foreground proposal, others could be arbitrarily assigned as foreground or background. 
This complexity of the evaluation process is already so overwhelming, not to mention the maximization, that numerous assumptions have been adopted in the literature to simplify it. For instance, \\
\textbf{1)}. A trivial reduction method is utilizing the spatial smoothness prior to reduce the effective $N$ to $N' (N' \ll N)$, as generated bounding boxes with relatively large IoU (typically larger than 0.5) are considered to belong to the same class; \\
\textbf{2)}. Some work in weakly-supervised learning assume there is only one instance for each label in the image \cite{bilen2016weakly}, which reduces the complexity to $\mathcal{O}(N')$ for each foreground class; \\
\textbf{3)}. The summation in Eq. \ref{eq:Q2} can be approximated using only its maximum value,
\begin{equation}
\label{eq:max_q}
Q(c=1, \bI) \approx \max_{\bt \in \mathcal{B}}P(\bt | \bI; c; \btheta')\log P(\bt | \bI; \btheta),
\end{equation}
yet one still needs to find the proposal assignment protocol that maximize this equation; \\
\textbf{4)}. The model output $\bt$ that maximizes $P(\bt | \bI; c; \btheta')$ are assumed to maximize $P(\bt | \bI; c; \btheta')\log P(\bt | \bI; \btheta)$ in Eq.\ref{eq:max_q} and is then used to calculate $Q(c=1, \bI)$. \\
\textbf{5)}. Preset a hard score threshold $p_t$, and assume that $P(\bt | \bI; c; \btheta')=\prod_{m \in \mathcal{A}_P} P(\tilde{\bt}_{s(m)} | \ba_m)\prod_{m \in \mathcal{A}_N}{P(c=0 |\ba_m)}$ is the maximum when $\mathcal{A}_P$ and $\mathcal{A}_N$ are the set of positive and negative proposals separated by a predefined confidence threshold $p_t$ (\emph{e.g.} 0.9). 

After these 5 steps of simplifications, the original task with complexity $\sim O(2^N-1)$ reduces to one $\sim O(1)$. These flow of assumptions and simplification are commonly seen in state-of-the-art semi- or weakly-supervised learnings. For example, only one $\bt$ is chosen out of the $2^N$ all possible proposal assignments and a hard confidence threshold ($\emph{e.g.}$, $0.9$ in \cite{sohn2020simple} and $0.6$ in \cite{nguyen2019semi}) is chosen to differentiate the foreground and background pseudo labels.



\textbf{Random pseudo-label sampling (RPS)}. Many of the above assumptions when reducing the dimensionality of $Q(c=1, \bI)$ in Eq. \ref{eq:Q2} are indeed just desperate choices and are far from being valid. In this study, we propose a random pseudo-label sampling (RPS) strategy, which can directly bypass the assumption chains and proves able to bring significant performance improvements.

\begin{equation}
\label{eq:sample}
\begin{split}
Q(c, \bI) &= \sum_{\bt \in \mathcal{B}}P(\bt | \bI; c; \btheta')\log P(\bt | \bI; \btheta) \\
          &=\mathbb{E}_{\bt \sim P(\bt | \bI; c; \btheta')}\left[ \log P(\bt | \bI; \btheta)\right] \\
          &\approx \frac{1}{B'} \sum_{\bt \in \mathcal{B}'} \log P(\bt | \bI; \btheta),
\end{split}
\end{equation}
where $\mathcal{B}'$ is the sampled set of $\bt$ according to $\bt \sim P(\bt | \bI; c; \btheta')$, $B'$ is the length of the sampled set $\mathcal{B}'$. 
Eq. \ref{eq:sample} basically transforms the summation over $2^N-1$ terms into $B'$ ones. And in this study we simply choose $B'=1$, meaning only one $\bt$ (\textbf{not} $\tilde{\bt}$) is sampled at each iteration. Since negative proposals are overall simple prediction and account for the vast majority of predictions\cite{pang2019libra, lin2017focal}, the set $\mathcal{B}'$ that consists at least one positive proposals is actually confined to a small space in $\mathcal{B}$. Therefore, reducing $B'$ to a small number or even $B'=1$ does not bring much variance during the estimation. 

The sampling process goes as Algorithm \ref{alg:rps}.
It involves two sampling processes. When the pseudo boxes are filtered using nms, conventional output only contains the most-likely representative at each location. Here we use a \textit{nms\_group} operator to output a set of index groups. In each group, all indices that are originally filtered by nms are kept in descending order w.r.t their scores. This is to guarantee that pseudo-boxes having a relative overlap can also be sampled, an approach to relax Assumption (1). 
The first boxes at all nms groups now form the originally nms-filtered pseudo boxes. Therefore, Step 1 in L-\ref{alg:step1} is to sample at different locations and Step 2 in L-\ref{alg:step2} is to sample inside a nms group. In this regards, every possible combination of pseudo-boxes from each proposals is possible in theory, relaxing Assumptions (1-5) directly.  

\begin{algorithm}[t]
	\caption{Random pseudo-label sampling (RPS)}
	\label{alg:rps}
	\begin{algorithmic}[1]
		\footnotesize
		\STATE {\bfseries Input:} Unlabeled image $\bI$, image-level labels $\bc$, agent model $\btheta'$
		\STATE {\bfseries Output:} Sampled pseudo label sets $\vect{X}'$, $\vect{S}'$
		\STATE Inference $\bI$ with $\btheta'$ yields the predicted bboxes $\vect{X}$ and classifications scores $\vect{S}$
		\FOR{$k = 1$ \TO $C$}
		\STATE \textbf{if} $\bc_k = 0$ \textbf{then}
		\STATE \ \ \ \ \ continue \COMMENT{Skip BG category}
		\STATE $\bG \leftarrow $ \textit{nms\_group}$(\vect{X}, \bS)$
		
		\COMMENT{Initialize the pseude label set $\vect{X}'$ and $\bS'$}
		\STATE $\vect{X}' \leftarrow \emptyset$, $\bS' \leftarrow \emptyset$
		\FOR{$g$ \textbf{in} $\bG$}
		
		\STATE \COMMENT{Step 1: Keep/drop group by its maximum score}
		\STATE \textbf{if} $p \sim \mathcal{U}(0,1) > max(S_g)$ \textbf{then}
		\label{alg:step1}
		\STATE \ \ \ \ \ continue
		
		\STATE \COMMENT{Step 2: Draw a sample from the group by its score}
		\STATE $\bS_g \leftarrow \{s_i \in \bS: i \in g\}$
		\label{alg:step2}
		\STATE $\bS_g \leftarrow \bS_g / sum(\bS_g)$ \COMMENT{normalize to [0, 1]}
		\STATE $i \leftarrow$ \textit{random\_choice}$(g, p=\bS_g)$
		\STATE $\vect{X}' \leftarrow \vect{X}' \cup \vect{X}'_i$, $\bS' \leftarrow \bS' \cup \bS'_i$
		\ENDFOR
		\ENDFOR
		\STATE \RETURN $\vect{X}'$, $\bS'$
		\STATE \R{\textit{nms\_group($\vect{X}$, $\bS$)}} return a group set $\bG$ where each element $g \in \bG$ is an index group in descending order of scores and represents the indexes of bboxes that are filtered using NMS.
		\STATE \R{\textit{random\_choice($g$, $p$)}} return a sampled index $i$ that draw from the index pool $g$, $p$ gives the chance that index to be picked.
	\end{algorithmic}
\end{algorithm}

Note that even though there is only one proposal assignment strategy after both Assumption (5) and Eq. \ref{eq:sample}, their theoretical foundations differ significantly.
Assumption (5) uses an empirical score threshold to differentiate the true pseudo-labels from the background, which is deterministic for each image in every step. This involves meticulous adjustment to find the best threshold value. Moreover, choosing a hard threshold brings unavoidable confirmation bias \cite{arazo2020pseudo}. That is, less confident foreground instances will never be chosen and over confident false positives will be given too large credit, which causes a bias for the target detector in Stage2. 
Whereas in RPS, the sampling method enlarges the space of pseudo-labels. All instances at different locations of different confidence scores, or pseudo-boxes inside one NMS group (the group that contains all boxes filtered during NMS) could be sampled according to their confidence scores. This has an combining effect of soft pseudo-labels and psedo-label voting. Moreover, the sampling process reduce the reliance of the target detector on the agent, reducing the possible confirmation bias and increase the generalization ability of the target detector.

\section{Experiments}
\label{sec:experiments}

\begin{table}
  \centering
      \begin{tabular}{l| c  }
      \hline  
      \textbf{Supervised}        & mAP(\%)  \\
      \hline
      Faster R-CNN (VOC07)     & 73.15     \\
      \hline
      Faster R-CNN (VOC07, 2x)           & 72.13 \\
      \hline
      Faster R-CNN (VOC07+12, 2x)           & 79.79 \\
      \hline
      SSD512 (VOC07)\cite{jeong2020interpolation} & 73.30 \\ 
      \hline
      \hline
      \textbf{Semi-supervised}           & mAP(\%) \\
      \hline
      SSD512+CSD\cite{jeong2019consistency} & 75.80 \\
      \hline
      SSD512+CSD+ISD\cite{jeong2020interpolation} & 76.77 \\
      \hline
      STAC \cite{sohn2020simple} & 76.77 \\
      \hline
      WSSOD (ours) & 78.00 \\
      \hline 
      \hline
      \textbf{ Weakly- and Semi-supervised}           & mAP(\%) \\
      \hline
      WSSOD (ours)& 78.90 \\
      \hline
      \end{tabular}
  \caption{
    Comparison on mAPs for PASCAL VOC dataset, where VOC07 is fully-labeled set and VOC12 is weakly-labeled set. 
    The number of iterations for Supervised and Supervised 2x setting on Faster R-CNN is 90k, 180k respectively.
    For results of SSD512, we directly borrowed them from \cite{jeong2020interpolation, jeong2019consistency}.
    For STAC, we implement their methods and re-trained it, since they reports the performce under COCO evaluation in paper.
    }
  \label{table: wssod on voc}
\end{table}

\begin{table*}[!t]
    \centering
        \begin{tabular}{l|| c |c| c |c |c }
        \hline
        Methods    & 1\% COCO & 5\% COCO & 10\% COCO  & 20\% COCO & 30\% COCO  \\
        \hline
        Supervised                & 10.6 & 19.2 & 24.1 & 29.4 & 33.6 \\
        \hline
        Supervised (2x)           & 9.0 & 18.0 & 23.1 & 27.6 & 31.8 \\
        \hline
        \hline
        WSSOD-Target (ours)              & 18.4 & 27.4 & 31.3 & 35.0 & 36.1 \\
        \hline
        \end{tabular}
    \caption{Comparison on mAPs for different ratio of fully-labeled data, where 
    Stage1 refers to the agent detector and Stage2 refers to the target detector. 
    Results are evaluated on COCO test dev.}
    \label{table: wssod on coco}
\end{table*}

To demonstrate the effectiveness of WSSOD, we conduct experiments on MS-COCO\cite{lin2014microsoft} 
and PASCAL VOC\cite{everingham2010pascal} datasets, which are the most popular datasets in object detection. 
MS-COCO is consists of 115,000 trainval images of 80 categories, while VOC07 and VOC12 contains 5,011 and
 11,540 images belonging to 20 classes respectively. 
Our experiments are conducted based on two settings. 
For MS-COCO, similar to STAC\cite{sohn2020simple}, we randomly sample 1, 5, 10, 20 and 30\% data as 
fully-annotated set and take the rest as weakly-labeled set. 
The corresponding results are evaluated on MS-COCO test-dev set. 
As for PASCAL VOC, we utilize VOC07 as fully-annotated set and VOC12 as weakly-annotated set, the evaluation results
are reported based on VOC07 test-dev set.

\subsection{Implementation Details}
Though WSSOD doesn't restrict the agent and target detector to be the same type, 
we choose the typical Faster RCNN\cite{ren2015faster} for both of them for simplicity.
The codes used for our experiments are based on Pytorch.
For MS-COCO experiments, we basically follow the \textbf{quick} training schedule proposed in STAC\cite{sohn2020simple},
which trains the agent detector for 90k iterations (1x) and target detector for 180k (2x) iterations
\footnote{STAC trained Stage1 and Stage2 for both 180k iterations but with the batch size of 8 and 16 respectively.
For simplify, we double the batch size but halve the number of iterations for Stage1.}.
As for PASCAL VOC, we train the agent detector and target detector for 120k, 240k iterations respectively with batch size 16.
In our setting, both fully-annotated and weakly-annotated samples sit together with equal amounts in a mini-batch.
As for the strong augmentation $T$ in Stage2, we simply uses the same strategy in STAC\cite{sohn2020simple}. 
It is extended from the RandAugment\cite{cubuk2019autoaugment} and contains transformations on color, 
global geometry, box-level geometry and cutout\cite{devries2017improved}.
For the loss weight of weakly-annotated data in Stage2, we take $\lambda_u=2$ since it has been testified to be optimum in \cite{sohn2020simple}.

\subsection{Results}

\label{results}

Though semi-supervised and weakly-supervised object detection have been widely studied,
rare works are related to weakly- and semi-supervised setting.
Consequently, we mainly compare WSSOD with semi-supervised methods and fully-supervised settings.
The experiments on PASCAL VOC are carried out by choosing VOC07 train/val dev as fully lableed set and VOC12 train/val dev as weakly-labeled set.
We reports the evaluation results on VOC test dev, which is shown in Table \ref{table: wssod on voc}.
This table is composed of three parts: Firstly, we show the results under fully-supervised setting. 
For Faster R-CNN, we trained for both 90k and 180k (2x) iterations to make fair comparison to WSSOD, since WSSOD trains 180k iterations in Stage2.
However, the longer training schedule only brings negative effectiveness under fully-supervised setting, which may be contributiond to the overfitting problem on small scale dataset.
We also reports the performance of SSD512 by directly borrowing the results from \cite{jeong2019consistency, jeong2020interpolation},
whose performance is better than Faster R-CNN.
Secondly, we exhibit the results under semi-supervised setting. 
For fair comparison, we omit the category label and abandon Global Classification Loss and Label Attention in Stage1,
which finally yields an agent model of 73.15\% mAP. 
With the proposed RPS module, we obtain a target model of 78.0\% mAP based on Faster R-CNN, which outperforms \cite{jeong2019consistency, jeong2020interpolation}, though they are trained based on a stronger detector.
Also, we beat the pseudo label based method STAC\cite{sohn2020simple}, which can demonstrate the effectiveness of the proposed RPS.
Thirdly, we demonstrate the result of WSSOD under weakly- and semi-supervised setting. 
With the aid of Global Classification Loss and Label Attention, we boost the performance of agent model from 73.15\% to 74.59\%, and further obtained a target model of 78.90\% mAP.
It is worth mentioning that the performance of Faster R-CNN under strongly-supervised setting (trained with VOC07 + VOC12) is 79.79\% mAP.
It means we achieve a performance that is comparable to fully-supervised setting with only about one third (VOC07) fully-labeled data. 


The experiments on COCO are carried out by randomly taking part (i.e. 1, 5, 10, 20 and 30\%) of its 
trainval dev as fully-labeled set and using the rest as weakly-labeled set. 
For fair comparison, we also trained model under fully-supervised setting, which only considers the 
available fully-annotated data to tain.
The evaluation results of Stage1 and Stage2 are reported in Table \ref{table: wssod on coco}.
We design the Supervised setting as contrast together to agent detector and Supervised (2x) as constrasts to target detector for fairness.
Supervised setting trains the detector with 90k iterations with batch size of 16, Supervised (2x) increase the number of iterations to 180k.
As we can see, using 2x training schedule leads to poorer performance due to the overfitting problem.
The performance of agent detector exceed all supervised setting with the aid of weakly-annotated data and label attention.
For example, it reaches 28.9\% mAP under 10\% protocol, which is 4.8\% higher than the supervised setting of equal training iterations.
However, a stronger agent detector is only half success, a stronger target detector is the ultimate aim.%
For Stage2, our methods reaches 36.1\% mAP under 30\% protocol, which almost reaches the performance under completely fully-supervised (i.e. 100\% COCO) setting, 37.6\% mAP.

\subsection{Abalation Study}

\subsubsection{Stage1: Global Classification Loss and Label Attention}

\begin{table}
  \centering
      \begin{tabular}{c | c | c  }
      \hline  
      Global Loss  & Label Atten. & mAP(\%)  \\
      \hline
       & & 24.1 \\
       \checkmark &  & 26.9 \\
       & \checkmark & 26.2 \\
       \checkmark & \checkmark & 28.9 \\
      \hline
      \end{tabular}
  \caption{Effects of Global Classification Loss and Label Attention on agent model performance under 10\% COCO protocol.}
  \label{table: abalation on agent}

\end{table}

\begin{table}[!t]
  \centering
  \begin{tabular}{l | c | c c c | c}
    \hline
    \multicolumn{2}{c|}{Methods} & 1\% COCO& VOC \\
    \hline
    \multirow{2}{*}{Semi Supervised} & $\tau=0.9$ & 15.2\%   & 76.77\%  \\
                                     & RPS & 15.6\%  &  78.00\% \\
    \hline
    \multirow{2}{*}{Weakly- and Semi-} & $\tau=0.9$ & 18.2\%  & 77.52\% \\
                                       & RPS & 18.4\% & 78.90\%\\
    \hline
  \end{tabular}
  \caption{Given the same agent model, the performance of target model with two different methods of generating pseudo label on COCO and VOC dataset.}
  \label{table: thr}
\end{table}
In Sec. \ref{sec:methodology}, Global Classification Loss and Label Attention Module were presented to obtain a stronger agent model.
Here we separately evaluate their influence on the agent model under 10\% COCO protocol, as shown in Table \ref{table: abalation on agent}.
Using Global Classification along, an improvement of 2.8\% mAP is obtained on agent detector, which demonstrates the effectiveness of leveraging weakly-labeled data in Stage1 training.
Separately employing Label Attention Module also brings 2.1\% mAP improvements.
This can be attributed to the reason that label attention module extracts the context information behind category labels.
This insight is close to \cite{hu2018relation}, which considers the spatial relation and category of proposals in object detection.
We argue that Global Classification Loss and Label Attention Module are proposed for different intuitions, they can benefit from each other.
By jointly using both of them, we finally achieved an improvements of 4.8\% mAP on agent model.

\subsubsection{Stage2: Random Pseudo-Label Sampler}

In this part, we conduct experiments under both semi-supervised and weakly- and semi-supervised to demonstrate the effectiveness of the proposed Random Pseudo-Label Sampler.
Since the way to generate pseudo label only influence Stage2, we take the same agent detector but using two different methods to generate pseudo label.
The most widely used method for choosing pseudo label is hrad-threshold \cite{sohn2020simple,sohn2020fixmatch, chen2020temporal}, as described by Simplification 5.
To be more detailed, a fixed threshold $\tau$ is firstly given, and used to drop those predictions with probability lower than $\tau$.
Previous works \cite{sohn2020simple} have done detailed research on the optimum threshold $\tau$.
In order to avoid repetitive labors, we directly borrow the conclusion of \cite{sohn2020simple} and use the optimum $\tau=0.9$ in both VOC and COCO for contrasts.
Experiments are conducted on both PASCAL VOC and MS-COCO dataset, as shown in Table \ref{table: thr}.
It's obvious that using RPS module yields better target model, which means  pseudo label with higher quality were generated.
In additional, our RPS module is parameter-free and doesn't need to tune for different cases.

\section{Conclusion}
In this study, a pipeline for weakly- and semi-supervised object detection (WSSOD) is proposed. WSSOD utilizes a small portion of full annotations with bounding box information and a large quantity of weak annotations with only image-level multi-labels. It is able to achieve a better balance between the labeling labor and the detector's performance. WSSOD is composed of two stages with the first stage training an agent detector and the second stage training a target detector using the pseudo label generated by the agent. Starting from a unified EM framework, the current pipeline is also carefully examined and improved. The weakly-supervised loss and label attention module are adopted during Stage1 training, to achieve a better agent model and thus a more accurate posterior estimate of the hidden bounding boxes in weakly-supervised data. The traditional pseudo-label generation process are then demonstrated to involve a chain of poorly-founded assumptions at Stage2. A random pseudo-label sampling (RPS) module is then proposed to bypass the chain of assumptions and directly sample a group of pseudo-labels according to the probability. RPS is more theoretically grounded and proves to be effective in both semi-supervised and WSSOD settings. With the above improvements, the proposed WSSOD is able to achieve $78.9\%$ AP on Pascal-VOC12 test set, only $0.9\%$ AP behind the fully-annotated baseline. The efficacy of WSSOD is also verified on MS-COCO dataset.

{\small
\bibliographystyle{ieee_fullname}
\bibliography{egbib}
}

\end{document}